\documentclass[letterpaper, 10 pt, conference]{ieeeconf}  

\IEEEoverridecommandlockouts                              
\usepackage[utf8]{inputenc}
\usepackage[T1]{fontenc}
\usepackage{graphicx}
\usepackage{url}

\usepackage[breaklinks]{hyperref}

\usepackage{xcolor}
\usepackage{tabularx}
\usepackage{tablefootnote}
\usepackage{textcomp,gensymb}

\title{\LARGE \bf
Nano Drone-based Indoor Crime Scene Analysis*
}

\author{Martin Cooney,$^{1}$ Sivadinesh Ponrajan,$^{1}$ and Fernando Alonso-Fernandez$^{1}$
\thanks{*We gratefully acknowledge support from the Swedish Innovation Agency (Vinnova) for the project "AI-Powered Crime Scene Analysis" and advice from the Swedish Police}
\thanks{$^{1}$M. Cooney, S. Ponrajan, and F. Alonso-Fernandez are with the School of Information Technology, Halmstad University
301 18 Halmstad, Sweden
        {\tt\small martin.daniel.cooney@gmail.com}}%
}

\begin{document}

\maketitle
\thispagestyle{empty}
\pagestyle{empty}

\begin{abstract}
Technologies such as robotics, Artificial Intelligence (AI), and Computer Vision (CV) can be applied to crime scene analysis (CSA) to help protect lives, facilitate justice, and deter crime, but an overview of the tasks that can be automated has been lacking.
Here we follow a \emph{speculative prototyping} approach:
First, the \emph{STAIR} tool is used to rapidly review the literature and identify tasks that seem to have not received much attention, like accessing crime scenes through a window, mapping/gathering evidence, and analyzing blood smears.
Secondly, we present a prototype of a small drone that implements these three tasks with 75\%, 85\%, and 80\% performance, to perform a minimal analysis of an indoor crime scene.
Lessons learned are reported, toward guiding next work.
\end{abstract}

\section{INTRODUCTION}
\label{section:intro}

The current paper reports on gaps and lessons learned designing a prototype of a small drone for indoor crime scene analysis.

Automation of crime scene analysis (CSA) is an important problem:
For example, violence alone was estimated to cost 14.76 USD trillion globally in 2017~\cite{iqbal2021estimating}.
When crimes occur behind closed doors, investigations can be required. 
Currently, however, investigations can be hindered by danger, human error, decomposition, and use of old-fashioned methods~\cite{swedishPolice2024}:
Investigators can suffer physical danger from criminals or traps at the scene, illnesses from contagious or toxic materials, and psychological harm such as vicarious trauma~\cite{physicalharm,psychologicaltrauma}. 
Victims as well often do not receive justice~\cite{salam2023underfunding}, due to various causes: 
\begin{itemize}
\item{{\bf Human Causes} include bias~\cite{gehl2017strategic}, errors and contamination~\cite{settingCrimeScenePerimeters}, as well as understaffing and underfunding~\cite{salam2023underfunding}.}
\item{{\bf Natural Causes} include degradation of evidence over time~\cite{benson2003without}.}
\item{{\bf Old-fashioned Methods}, which could introduce inefficiencies or inaccuracies, include measuring bloodstains by hand using using rulers, strings, and protractors~\cite{oldblood}, sketching using paper, pencils, erasers, rulers, and clipboards~\cite{oldsketching}, or carrying powerful light sources and filters (which can weigh ten kilograms) to work in dark rooms (e.g., ~\cite{heavyexample1,heavyexample2}).}
\end{itemize}

\begin{figure}
\centering
\includegraphics[width=.5\textwidth]{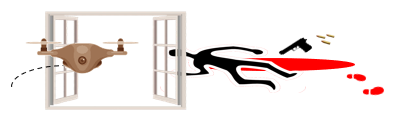}
\caption{Basic concept: a drone could help to quickly and safely analyze crime scenes, (a) accessing the scene to gain situation awareness, (b) gathering evidence, and (c) conducting initial analysis for an investigation.} \label{fig_intro}
\end{figure}

As in Fig.~\ref{fig_intro}, we imagine that small drones which can navigate in tight indoor spaces could help investigators and victims by accessing crime scenes quickly, reducing risks of contamination, and using sensory and visualizing modalities not available to humans to better capture and share information. 
Thereby, the idea is not that humans should be replaced, but that a drone can help humans by providing rapid initial processing.

Here, the term "drone" is defined loosely to comprise unmanned aerial vehicles (UAV), unmanned aircraft systems (UAS), unmanned aircraft vehicle systems (UAVS), remotely piloted aerial vehicles (RPAV), and remotely piloted aircraft systems (RPAS), including quadcopters, quadrocopters, quadrotors, (ultra-lightweight/tiny) micro aerial vehicles (MAVs), and "pocket drones".
Furthermore, we follow the convention that "nano drones" are less than 250 g~\cite{dean2024nano}.
Various nano drones have been created for research, commercial, or military use, that could be adapted for CSA, as shown in Table~\ref{tabNanoDrones}.

\begin{table}
\caption{Some nano drones that could be used for indoor CSA.}
\label{tabNanoDrones}  
\begin{tabularx}{\linewidth}{ | >{\hsize=0.75\hsize}X | >{\hsize=0.25\hsize}X | } \hline
Harvard RoboBee~\cite{helbling2018altitude} 
& 0.08 g  \\ \hline
Fünfiiber~\cite{muller2021funfiiber} 
& 18 g  \\ \hline
AeroVironment Nano Hummingbird~\cite{hummingbird} 
& \~19 g \\ \hline
DelFly Explorer~\cite{de2014autonomous} 
& 20 g \\ \hline
Blade Nano QX~\cite{bladenano} 
& 22 g \\ \hline
Crazyflie~\cite{crazyflie} 
& 27 g \\ \hline
(Custom) Blade mCX2 Helicopter~\cite{moore2014autonomous,helicopter} 
& 30 g \\ \hline
Teledyne FLIR Black Hornet 3/4~\cite{hornet3,hornet4}
& 32 g/70 g   \\ \hline
(Edge-FS Pocket drone)~\cite{mcguire2017efficient}
& 40 g   \\ \hline
(46g quadrotor)~\cite{briod2013optic}
& 46 g   \\ \hline
Trashcan drone~\cite{li2020visual} 
& 72 g  \\ \hline   
DJI Ryze Tello~\cite{tello}  
& 80 g  \\ \hline
\end{tabularx} 
\end{table}

A challenge is that it's unclear which tasks nano drones could do, since CSA is highly complex. To gain insight, we follow a \emph{speculative prototyping} approach~\cite{tironi2018speculative}, guided by the \emph{STAIR} "tool" from forensic science~\cite{gehl2017strategic,stepsOfCSI}.
Speculative prototyping seeks to rapidly, openly, creatively and critically craft glimpses from possible future scenarios, starting with the question: "What if drones could help with CSA?"
The STAIR tool models how investigators think and structure their investigations in a systematic, organized manner, which is important given the complexity and multiplicity of processes to be carried out. It embarks from the idea that an analyst must understand the \emph{Situation}, carry out \emph{Tasks}, \emph{Analyze} evidence, and \emph{Investigate}, to obtain \emph{Results}~\cite{gehl2017strategic,stepsOfCSI}. Within this context, our contribution here is two-fold:
\begin{itemize}
\item{{\bf Theoretical}. We \emph{speculate} on how drones could be useful for CSA, including a list of some potential desired capabilities.}
\item{{\bf Practical}. We \emph{prototype} some capabilities for drones within a mock-up crime scene.}
\end{itemize}

Below, in Section~\ref{section:review}, we position the paper in regard to previous work, identifying gaps. Then, in Section~\ref{section:method}, we explore three gaps via prototyping, with evaluation results discussed in Section~\ref{section:discussion}. Our aim is to provide insights that could inspire discussion and interest, toward bringing justice in a faster, more accurate way.

\section{RELATED WORK}
\label{section:review}

To survey the area, top results were analyzed from ACM, IEEE Xplore, and Google Scholar with the search
phrase "(drone OR UAV OR robot) AND (crime)". Bibliographies of relevant papers were scoured, and some known papers also added. To try to avoid missing important information, we also followed up with some extra queries to Scite.AI and ChatGPT. Results outside of the scope of the paper were removed, regarding crimes by drones, security of drones, and forensic analysis of drones to catch human criminals.
This led to 32 papers being reviewed (newest: 2024, oldest: 1991), which are summarized below following the categorization scheme of the STAIR tool.

\subsection{Situation Awareness}
\label{subsection:drone_vision_crime}

The Situation step in STAIR involves accessing a crime scene to infer if the threat has passed and what has happened (where and when, and to whom).

Since the first "Drone as First Responder" (DFR) program started in 2018 in Chula Vista, California, drones like the Matrice 300 RTK have been remotely piloted thousands of times (e.g., from rooftops, like a mini-helicopter) to rapidly respond to 911 calls and provide enhanced situation awareness about outdoor incidents such as car accidents, vandalism, overdoses, and noise disturbances~\cite{mehrotra2024age}. (Such programs have now also spread to other cities in the US such as New Orleans and New York, as well as to other countries such as Sweden, indicating growing interest~\cite{polisen2024se}.)
Robots can furthermore detect doors and windows~\cite{mohamed2023yolo}, and enter buildings by opening doors~\cite{kcal2024look}, breaking walls or being thrown inside by a human~\cite{glaser2016eleven}.
Drones can even acrobatically pass through narrow gaps~\cite{falanga2017aggressive}. 
Some drones capable of threat detection have also been designed to detect dangerous traps, pathogens, and criminals, such as Improvised Explosive Devices (IEDs) with radar, disease-carrying animal carcasses via thermal camera~\cite{rietz2023drone}, and escaping or hiding people~\cite{zheng2019evolutionary,chia2022keeping}.
Offenses such as fighting have also been visually recognized~\cite{perez2019detection}.
More generally, robots have used observation to automatically infer the underlying meaning of human behaviors, in games like chasing, follow-the-leader, and tag~\cite{crick2010controlling}.

However, much remains to be done in this area. Our brainstorming suggested that gaps could include:
\begin{itemize}
\item{{\bf Access in challenging cases}. Drones could traverse partially open windows or doors, or pick locks.}
\item{{\bf Threat detection}. Criminals and victims hiding indoors (who might be paused or moving to avoid detection, e.g. in a closet, under furniture, or behind a wall in a neighboring room) could be detected via radar or thermal cameras. Microphones could also capture sounds from voices, gunshots, or breaking glass, to detect shooter locations or fleeing suspects.}
\item{{\bf Offense detection}. Foundation models and Large language models (LLMs) could be shown multimodal data (e.g., surveillance footage of crimes) and queried; i.e., to determine roles (even for groups coming and going), infer past/future actions, estimate levels of force, pinpoint focal points of a crime (location and cause, like where a person was stabbed or a fire or explosion started), detect anomalies (such as a broken lamp), and generate crime scene sketches, etc.}
\end{itemize}

\subsection{Tasks}
\label{subsection:tasks}

The Tasks step in STAIR involves dealing with active threats to ensure safety; freezing and controlling a perimeter; detecting, gathering, and preserving evidence; as well as cleaning.


Robots have been designed to deal with threats like IEDs~\cite{gwozdz2014enabling}, disinfect complex surfaces with ultraviolet irradiation~\cite{sanchez2024improving}, and safely extract samples from corpses
~\cite{neidhardt2022robotic}.
To avoid psychological harm to humans or privacy infringements, video can be altered via blurring, masking, or mosaics~\cite{jiang2024beyond}. 
Our previous work also lists various robots carrying weapons such as explosives, stun guns, pepper spray, water, firearms, or nets, some of which have been used to detect, negotiate with, or neutralize threats, including via computer vision~\cite{cooney2023broad}. 
A makeshift perimeter to protect evidence could also be created by dropping bricks at appropriate locations, a strategy which has been used previously by drones to construct a building~\cite{augugliaro2014flight}.

As shown in Table~\ref{tabEvidence}, previous work in securing evidence can be roughly split into manual piloting of a drone or use of a lidar to take 3D scans of a mock-up crime scene, toward assessing potential benefits or demerits of technology for CSA; variables of interest have mainly been mock-up bloodstains, guns, knives, and bodies.

\begin{table*}
\caption{Prior Work Related to Detecting and Scanning Evidence.}
\label{tabEvidence}  
\begin{tabularx}{\linewidth}{ | >{\hsize=0.2\hsize}X | >{\hsize=0.4\hsize}X | >{\hsize=0.4\hsize}X | } \hline
Urbanova et al.~\cite{urbanova2017using} & Drone/CV (3D scanning outdoors via a piloted DJI Phantom 2 drone with a GoPro HERO 4 and photogrammetry with Agisoft PhotoScan) & Target: Dummy, bones, and artificial blood, in grass (detected by humans)
\\ \hline
Georgiou et al.~\cite{georgiou2022uav} & Drone/CV (piloting a DJI Spark drone outdoors for accurate, fast, reliable detection, picking colors with Matlab) & 
Colored square foam pads
\\ \hline
Cooney et al.~\cite{cooney2024designing} & Drone/CV (piloting a DJI Ryze Tello drone indoors to detect via YOLO/heat traces) & Hidden cameras
\\ \hline
Bucknell and Bassindale~\cite{bucknell2017investigation} &
Drone (piloting a Parrot AR.Drone 2.0 at various heights to check the effects of downwash on sensitive evidence)
& Textile fibres on various substrates
\\ \hline
Rymansaib et al.~\cite{rymansaib2023prototype} & Drone/sonar  & A mannequin "body" (future target possibly also weapons, narcotics, or IEDs)
\\ \hline
Araujo et al.~\cite{araujo2019multi} & Simulation of a drone detecting evidence based on AirSim/Unreal Engine/YOLO & A body, bloodstain, gun, and knife (augmented MS-COCO data set)  
\\ \hline
Butt et al.~\cite{butt2023application}. & CV (YOLO)  & Bullet holes  
\\ \hline
Nandhini and Thinakaran~\cite{nandhini2023detection} & CV (classification via a 7-layer CNN) in infrared images & Firearms, knives, money, blood, animals, cars, and cellphones (although with some unclarity regarding datasets)  
\\ \hline
Buck et al.~\cite{buck2013accident} &
3D scanning & Bullet trajectories and damage to bodies and objects
 \\ \hline
Esposito et al.~\cite{esposito2023advances} & 3D scanning 
& Locations and distances between corpses, bloodstains, and weapons; bullet trajectories; paths where the victim might have moved before dying; and balcony height 
\\ \hline
Galanakis et al.~\cite{galanakis2021study} &
3D scanning & A "dead body" and various objects such as a phone, screws, tape, and a box (dataset released)
\\ \hline
Liscio et al.~\cite{liscio2020observations} & 3D scanning (using an apparatus) & "Cast-off" bloodstains--from blood flying off a weapon such as a bat, hammer, knife or pipe--estimating a "Path Volume Envelope" for the path the weapon took
\\ \hline
\end{tabularx} 
\end{table*}

Here, too, much remains to be done:
\begin{itemize}
\item{{\bf Handling threats}. A drone could estimate backdrops and infer where and when force could be used; as well, drones could clear rooms and enter in a fast, effective, unpredictable, and safe way, like SWAT teams.}
\item{{\bf Geofencing}. After detecting where a crime took place, a drone could create a perimeter, establish a "path of contamination" or "common approach path", and manage the scene--preventing unauthorized entry and improper actions.}
\item{{\bf Handling evidence}. 
When mapping, useful information, like the distances between samples (e.g., blood, bodies, and weapons) should also be calculated.
Another important step after detecting evidence is gathering (e.g., blood with a swab).
For blood, presumptive testing could be conducted with luminol or tetramethylbenzidine (Hemastix) mounted to a drone. 
Other important detection targets could include footprints/shoeprints, clothing (e.g., gloves, shoes), other weapons (e.g., bullets, bullet casings, gunshot/gunpowder residue, clubs), restraints (e.g., cable ties, handcuffs), extra DNA evidence (e.g., skin, hairs, body fluids other than blood--like saliva on cigarettes, drinks, food, or cutlery), and broken objects (e.g., glass/pottery fragments, paints, particulates, bullet holes, and fragment patterns (“fractography”)). 
Furthermore, any object close to evidence such as a corpse or blood might also be interesting.
Basic inference could be conducted on corpses (e.g., identity, pose, age, gender, and estimated time of death (based on body temperature)).
In addition to lidars, low-light/infrared/ultraviolet cameras, and gas sensors to detect bodies (e.g., via putrescine, cadaverine, methane, hydrogen sulphide, or ammonia), a thermal camera could help estimate when a person last touched evidence like weapons or drugs if a drone arrives fast enough.}
\item{{\bf Visualizing}. RViz in Robot Operating System (ROS), Unreal Engine (UE), or Unity, could be used to visualize, "replay" crimes, update models when new evidence is discovered, and prepare initial reports.
The data could be accessed also via Extended Reality (XR) or 3D printed, and different evidence could be marked with different colors.}
\item{{\bf Cleaning}. A robot such as Roomba could be used in simple cases~\cite{roomba}. More complex cases might require removing all carpet, scrubbing floors, and more intensive cleaning. Such a robot could be designed to be easy to wash and sterilize.}
\end{itemize}

 
\subsection{Analysis and Investigation}
\label{subsection:analysis}

In the Analysis step, theories can be drawn from raw data about motives, opportunities, and means. 
Theories from multiple sources (e.g., witness testimonies and physical evidence) are then compared in the Investigation step to narrow down what likely happened.

Various work explores blood pattern analysis (BPA):
For example, a review from Weber and Lednev describes how time since deposition (TSD) can be estimated using CV via magenta values, brightness, or blood pool/crack ratio, despite challenges due to effects of substrate, contamination, and environment (temperature and humidity)~\cite{weber2020crime}.
Of the three main kinds of bloodstains, some focus seemed to have been placed on the passive drips and active spatter, which have been, e.g., discriminated by Bergman et al. using Convolutional Neural Networks (CNNs)~\cite{bergman2022automatic}.
We could only find a few examples of transfer stains (smears and swipes) in one dataset, which included slow-motion videos of a person swiping clean or bloodied glass or cloth with a finger, cloth, paper towel, or toothbrush from below or from the side~\cite{bloodstain_transfer_dataset}. (Although insightful, extra work would be required to adapt such data for a drone intended to fly over evidence after a crime has been committed.)

As well, various mathematical models exist that could be used to infer how an object (such as a teapot) broke~\cite{norton1991animation}.
Robots could also be designed to publish a report summarizing detected anomalies, along with interviews from nearby people~\cite{matsumoto2007journalist}.
Furthermore, logic programming languages like Prolog, and inference approaches such as Maximum Likelihood Estimation (MLE) or its Bayesian counterpart Maximum a posteriori inference (MAP) could also be used to narrow down possibilities.

However, it seems little work has focused so far on the analysis and investigating steps for CSA robots, possibly since they build on the simpler preceding steps that have mostly not yet been implemented in robots. A few examples of specific gaps are as follows:
\begin{itemize}
\item{{\bf Transfer stains}. In some cases, blood patterns could carry information about how victims and criminals moved and so-called "second locations". Moreover, composite scenes comprising various kinds of patterns could be analyzed.} 
\item{{\bf Affordances}. The possibility for nearby objects to be used as makeshift weapons could be assessed when hunting for a missing weapon in a violent crime.}
\item{{\bf Motive Inference}. A missing or open safe could indicate financial motives--the opposite, for a murder victim with wallet intact. Also, a distinctive modus operandi could suggest the work of a serial criminal.}  
\end{itemize}

Thus, our review did not reveal a pre-existing overview of how drones could be used for CSA; however, it seemed that much inspiration could be taken from the literature, and that many opportunities exist for development, some of which we explore in the next section.

\section{METHOD}
\label{section:method}

Prototyping can reveal challenges and opportunities that might not be apparent if only theory is considered.
From the identified gaps, we first chose three initial capabilities that seemed feasible and useful to explore: accessing a crime scene through a partially opened window, mapping evidence, and analyzing motion from a bloody path.

\begin{figure}[t]
\centering
\includegraphics[width=.3\textwidth]{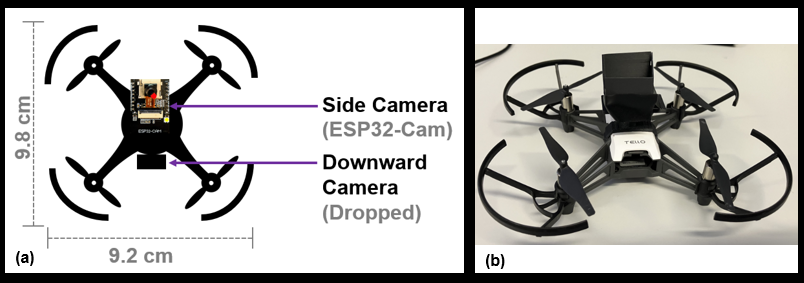}
\caption{Drone used for exploration, a modified Ryze Tello} \label{fig_tello}
\end{figure}

\begin{figure}[t]
\centering
\includegraphics[width=.5\textwidth]{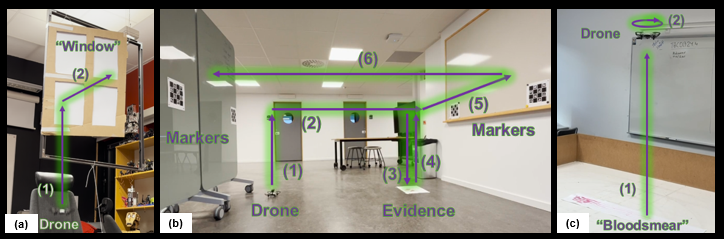}
\caption{Set-up for exploring each capability: (a) the drone rises and tries to move straight through a partially open "window", (b) the drone rises and moves between markers, pausing to "gather" detected evidence, and (c) the drone rises and rotates above a "blood smear", estimating direction.} \label{fig_experiment-setup}
\end{figure}

\begin{figure}[t]
\centering
\includegraphics[width=.5\textwidth]{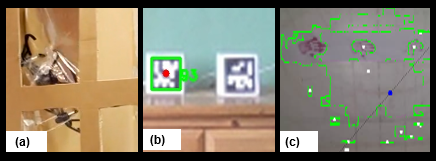}
\caption{Examples of initial challenges: (a) propellers getting caught, (b) single markers not being detected, and (c) false detections in low light} \label{fig_failures}
\end{figure}

\begin{figure}[h!]
\centering
\includegraphics[width=.5\textwidth]{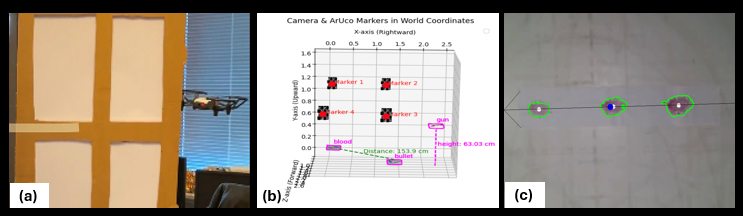}
\caption{Examples of successful trials: (a) pushing open a window, (b) mapping distances, and (c) inferring direction} \label{fig_successes}
\end{figure}

Then, we constructed proof-of-concepts.
For hardware, a DJI Ryze Tello drone~\cite{tello} was augmented to be able to see both below and in front, by popping out its side camera to point downwards, and adding a 3D printed stand on top of its chassis to carry an ESP32-Cam, powered by a lithium polymer battery, as shown in Fig.~\ref{fig_tello},
The Ryze Tello drone is inexpensive, easily programmed, and has a large payload (> 60 g); ESP32-Cam is also inexpensive and easily programmed to wirelessly stream SVGA@30fps from its 2 Megapixel OV2640 camera~\cite{esp32cam}. 
We also set up some software on an external laptop, including OpenCV~\cite{opencv} and YOLO 11~\cite{yolo}.
An overview of our set-up for exploring the three capabilities is shown in Fig.~\ref{fig_experiment-setup}. 
Also, Fig.~\ref{fig_failures} and Fig.~\ref{fig_successes} visually depict some of the challenges and successes encountered using this setup, which are detailed below. 

\subsection{Capability 1: Situation. Pushing Open a Window}
\label{subsection:window}

Rapid access is a crucial reason for using drones, and humans cannot be expected to be on scene to let them in, so drones could seek to gain entry in various ways, including via windows.
In some cases, a window might be partially open, such that acrobatic entry could be difficult, and lock-picking unnecessary, but how to handle such cases was unclear.

Yeong et al. (on page 4082, section IIIA, paragraph 3, line 8) estimated that opening a door requires 35 N~\cite{yeong2009reachman}.
Assuming similar forces required for windows and doors, a drone like the Drone Racing League (DRL) Racer3 with 71 N of thrust--possessing a hard shell to protect from crashes, weighing over 1kg and capable of accelerating to 36m/s in under a second--could potentially push open some partially opened windows~\cite{drl}.
However, safety is also a concern, since an error at high acceleration could lead to injuries.
This suggested the usefulness of starting out with a nano drone  for safely testing ideas.

Although hardware is continually advancing, it could be difficult currently for a nano drone to force open a tough window.
Based on Newton's second law of motion (\emph{F}=m\emph{a}), a small drone with approximate mass 100 g and acceleration 5m/s$^2$ would have only 0.5 N of thrust force.
Thus, we chose to prepare an initial mock-up window using a cardboard box with plastic wrap in place of glass.
Of the four basic kinds of window--rotating horizontally (casement) or vertically (awning), and translating horizontally (sliding) or vertically (double hung)--a common casement form was selected.
Various opening mechanisms for the drone were also considered, from wedges to actuated tongs.
Eventually, for initial testing, a plastic spike was attached to the underside of the drone with putty.

Initial trials with manually piloting the drone to open the window with the spike indicated substantial challenges with low force, downwash (inaccurate control), and propellers getting caught despite protective guards, with a performance of roughly only 20\%.
Therefore, we redesigned the test window to be lighter (91 g), without deep walls (string was used to hang it from the ceiling and pull the bottom down via a weight on the floor), and with paper in place of plastic wrap.
The revised set-up was checked by conducting 20 trials.
The result was 75\% successful performance, with the drone pushing through the partially opened window 15 times.
A main factor in the five failed attempts was the imperfect control due to the inexpensive drone used: for example, the drone sometimes came into contact with the window near its hinges where torque would be insufficient or was knocked outside of the area of the window.

\subsection{Capability 2: Tasks. Mapping Evidence}
\label{subsection:mapping-evidence}

After gaining access, a drone should map the crime scene, calculating distances between samples and gathering evidence, which was unclear how to realize.
To simplify our initial exploration, we set the drone to fly twice, once to create a 3D map, and the second time to find evidence. Also, a small 2 m x 2 m mock-up murder crime scene was created using some printed photos of a gun, blood, and bullet casings on A3 paper (29.7 cm width x 42.0 cm height), and four ArUco markers to mark walls. 
The photo of a gun was placed on a table, and the other photos on the ground.

First, a 3D map was initialized.
The Ryze Tello drone took off and rotated to detect markers via its side camera (ESP32-Cam), whose video feed was streamed through WiFi to an external laptop.
The first marker detected was assigned to a location at the origin relative to the x and z axes and at a height along the y axis inferred based on the drone's height, given by its bottom infrared range sensor.
As the drone rotated, newly detected markers were sequentially localized with respect to the first marker.
Once mapping was complete, the drone landed.

Second, the drone took off again, looking for evidence within the 3D map, by moving between markers.
The drone's video feed from its downward facing camera was again relayed to the external laptop, which used YOLO 11 with a GPU to detect three classes of object (guns, blood, and bullet casings).
Visual servoing was conducted to center the drone over the detected evidence.
Then, information about evidence was added to the 3D map, based on the drone's estimated location and altitude.
Furthermore, the drone was set to land on the center of the detected evidence to mimic gathering a sample (e.g., harvesting blood via a swab on its underside).
Finally, when the drone's search ended and its map was completed, the distances between all detected objects was calculated and output.

Initial trials indicated problems with ArUco markers being too small to be detected when the drone was far or too big when the drone was close. Eventually, a 4x4 ChArUco Board was adopted, which combines a chess board pattern and multiple ArUco markers to enhance accuracy and handle partial occlusions.

To evaluate the system, 20 trials were conducted, which resulted in an object detection accuracy of 85.0\%.
Reasons for some inaccuracy included illumination and lack of contrast between the bloodstain photo and the floor, as well as limited amounts of training images.
The average discrepancy between actual distances and estimated distances was also found to be 2.4 cm (SD: 0.5 cm).
This error might have been due to challenges with the stability of the drone when capturing images, illumination, and the limited resolution of the drone's cameras (ESP32-Cam), as a lower resolution was used to avoid lag.
Additionally, we noted that YOLO worked at 24 fps, and trials took approximately three minutes.

As noted by Bucknell and Bassindale, however, a drone moving overhead can disturb trace evidence like fibers, hairs, or small particles~\cite{bucknell2017investigation}. We expect this effect to be strong here, given that the drone flies twice, and even lands on evidence to simulate gathering. Thus, to avoid damage, such an approach should only be used when the absence of trace evidence has been verified, or an investigator has decided that the risk is acceptable.

\subsection{Capability 3: Analysis. Blood Smear Inference}
\label{subsection:blood-smear-inference}

Once evidence has been documented, it should be analyzed.
For example, BPA on transfer stains such as smears and swipes could indicate where a criminal went or a body was moved (e.g., to delay or confuse investigators).

For initial exploration, since we could not find a dataset that would fit the designated context, a simplified dataset was created using red paint on white paper as a substitute for human blood: viz., carmine dye (E120, from cochineal insects), on four white A4 sheets per sample (21.0 cm width x 118.8 cm length).
Body parts, continuity, and direction were varied.
Body parts included a hand, shoe, and an amorphous blob.
Continuity involved if the body part was constantly in contact with the paper ("legato", like a corpse being dragged) or set down at regular intervals ("staccato", like a shoe). 
Specifically, we walked over the paper with a soaked shoe (European size 43) and crawled with one soaked hand (length: 19.5 cm) to mimic a criminal or victim fleeing; and, we dragged a soaked hand, and touched down or dragged an amorphous blob using a soaked tissue, to mimic a corpse being moved. 
In total, twenty samples were created: 
Our drone was programmed to automatically take off and rotate over each of the five samples, capturing video (320 x 240). 
Then, for each video, images were manually captured from four views: up, down, left, and right.

Next, a Python program was written to calculate the perceived direction of motion, using a simplified approach involving color picking and image moments.
The intuition was that the majority of the detected red color should be closer to the start position where there is more blood than the end position where little blood remains.
First, red color was detected via hue (two ranges), then contours.
Then a line was fitted to the centroids of contours.
Finally, the position of the centroid of centroids along the line was used to estimate the direction the bloody object was moved with dot product and an adjustment for the quadrant. 
A closest codebook vector/class label was selected, and the error between prediction and ground truth calculated.

Initial trials showed various challenges, such as many small false contours using standard hue values for red (possibly due to illumination and the dye used), lines not being fit for long "legato" smears with a single contour, as well as some errors estimating direction with multiple contours from assigning equal weight to large and tiny contours. 
Therefore, the range for red was adjusted (e.g., removing orange-red), contour area was checked, code was written to handle base cases (e.g., in cases with only one contour, which can happen if an object is being dragged, moments are used to find a line based on the contour's shape), and the midpoint of the detected blood pixels was used in place of the center of centroids to estimate direction.

As a result, accuracy was 80\%, with an average error of 50.3\degree.
Accuracy was imperfect due to a similar amount of red color being detected at the start and end for two samples: While hand staccato, shoe staccato, hand legato were completely detected, the amorphous "tissue" stains seemed to be predicted at the level of random chance (50\%), and we could also not tell from observation which side was the start or end point.

Likewise, the average error was large for the same reason, the amorphous tissue stains. 
Of the four error cases, three were off by 180\degree, since start and end points were reversed.
We feel that this error is not a crucial problem for the following reasons:
The "line of motion" is estimated correctly in 19 of 20 cases (95\% accuracy).
We don't expect all transfer stains at crime scenes to be amorphous tissue stains; the error for correct cases was 26.2\degree, which seems reasonable.
Even considering the challenging tissue stains, 50.3\degree is less than the error expected from a random guess, which would be 180\degree.
As well, our goal is only to explore one possible simplified approach to give an initial estimation; the approach can be improved, and the initial estimate later verified by humans or robots.
Thus, the results indicated that some cases can be easier to infer than others.

\section{DISCUSSION}
\label{section:discussion}

Imagining potential benefits to speed, accuracy, and safety, we presented an overview of indoor crime scene analysis (CSA) by a nano drone:
\begin{itemize}
\item{{\bf Theoretical}. First, we identified some theoretical gaps in eleven categories, based on a "big picture" (rapid scoping) review.}
\item{{\bf Practical}. Second, some practical challenges, summarized in Table~\ref{tabChallenges}, were explored by iteratively designing three prototype capabilities relating to accessing crime scenes through windows, mapping and gathering evidence, and analyzing blood smears, with performances of 75\%, 85\%, and 80\%.}
\end{itemize}
A summary video and some code were also made available~\cite{cooney2025nanoVideo,cooney2025nanoCode}.

\begin{table}
\caption{Some practical challenges identified.}
\label{tabChallenges}  
\begin{tabularx}{\linewidth}{ | >{\hsize=0.25\hsize}X | >{\hsize=0.75\hsize}X | } \hline
Gaining entry to a crime scene through partially opened windows
& Low thrust force, propellers getting caught despite protective guards, and inaccurate control (downwash, contacting the window at inefficient locations near the hinges, and being knocked away).  \\ \hline
Mapping and gathering evidence
& Low illumination and low image resolution, lack of contrast between objects to detect and the floor, limited amounts of training images, and drone instability \\ \hline
Estimating motion of bloody objects (i.e., the "dragged direction")
& Color picking in low illumination with standard hue values for red, detection of only one contour or many tiny contours, and amorphous stains from soaked tissues dispensing an equal amount of red at the start and end points.\\ \hline
\end{tabularx} 
\end{table}

In summary, while nano drones do not seem capable yet of, e.g., skillfully opening windows, gathering evidence, and analyzing complex blood patterns in the real world, we believe that our results suggest the feasibility of initial ideation and testing within a laboratory setting. 
Additionally, we have aimed to follow a sustainable approach--in using a low-cost, low-energy platform and documenting prototyping in an open way--toward supporting a positive societal impact.

\subsection{Limitations and Future Work}
\label{section:limitations}

Our approach is limited by its exploratory nature:
We expect many more important theoretical questions could exist that were not discussed here.
For example, wireless transmission of potentially sensitive data from drones to remote laptops for processing could present an security threat, allowing criminals to intercept and possibly alter communications between drones and humans.
As well, the scope of this paper did not allow us to explore all of the gaps we identified.
Even for the explored capabilities, we expect huge discrepancies between real-world scenarios and our exploratory experiments, involving rapid, low-fidelity prototyping; e.g., using mock-up windows, small lab spaces with markers, and photos or red paint in place of real firearms or human blood.

\begin{figure}
\centering
\includegraphics[width=.5\textwidth]{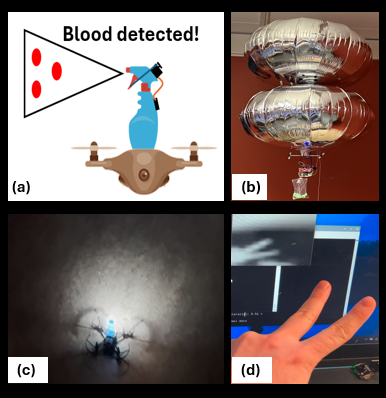}
\caption{Additional Prototyping: (a) Revealing hidden blood stains, (b) documenting evidence without disturbance via a drifting blimp, (c) using a flashlight in the dark, and (d) detecting hidden people with thermal or radar.} \label{fig_additional_prototypes}
\end{figure}

Ideation from researchers of various backgrounds will expose new important theory. 
Future work should also move farther from the laboratory, and into the real world, to better make conclusions about the applicability of the designated capabilities:
Stronger, more advanced drones will autonomously fly in more accurate settings, onboard sensors like lidars will be used rather than markers for mapping, and real blood stain patterns will be analyzed.
In-built processing will also enhance privacy (possibly along with federated learning and selective blurring); possible disturbances (i.e., destruction of sensitive evidence) due to a drone's static electricity, electromagnetic interference, lights, or heat should be thoroughly checked; and potential misuses by criminals identified (e.g., unlawful entry via windows for burglary, or detection and tracking of innocent citizens by stalkers).
Additionally, we have started to explore a broader range of the capabilities identified as potential gaps, conducting some development on each of our ideas. These additional scenarios are visualized in Fig.~\ref{fig_additional_prototypes}, and in the accompanying video.

In conclusion, while much work remains to be done to make this vision of nano drones helping to solve crimes a reality, we believe that the day is not too far off in which such technologies will help to preserve justice and otherwise impact society in a positive way. 

\color{black}

\addtolength{\textheight}{-0cm}   







\bibliographystyle{IEEEtran}
\bibliography{IEEEabrv,drone_crime}





\end{document}